\documentclass[10pt,twocolumn,letterpaper]{article}
\usepackage{amsmath}
\usepackage{amssymb}
\usepackage{graphicx}
\usepackage{cite}
\usepackage{url}
\usepackage{hyperref}
\usepackage{booktabs}
\usepackage{multirow}
\usepackage{array}
\usepackage{float}
\usepackage[margin=1in]{geometry}

\usepackage{titlesec}
\titlespacing*{\section}{0pt}{8pt plus 2pt minus 2pt}{6pt plus 2pt minus 2pt}
\titlespacing*{\subsection}{0pt}{6pt plus 2pt minus 2pt}{4pt plus 2pt minus 2pt}

\usepackage{enumitem}
\setlist{nosep,leftmargin=*,itemsep=0pt,parsep=0pt,topsep=2pt}

\usepackage{etoolbox}
\patchcmd{\thebibliography}{\setlength{\itemsep}{0pt plus 0.3ex}}{\setlength{\itemsep}{0pt}}{}{}

\title{ECG Classification on PTB-XL: A Data-Centric Approach with Simplified CNN-VAE}

\author{
    \textit{Naqcho Ali Mehdi}$^{1}$, \textit{Aamir Ali Drigh}$^{1}$ \\
    $^{1}$NED University of Engineering and Technology, Karachi, Pakistan \\
    Email: naqchoali@gmail.com
}

\date{}

\begin{document}

\maketitle

\begin{abstract}
Automated electrocardiogram (ECG) classification is essential for early detection of cardiovascular diseases. While recent approaches have increasingly relied on deep neural networks with complex architectures, we demonstrate that careful data preprocessing, class balancing, and a simplified convolutional neural network combined with variational autoencoder (CNN-VAE) architecture can achieve competitive performance with significantly reduced model complexity. Using the publicly available PTB-XL dataset, we achieve 87.01\% binary accuracy and 0.7454 weighted F1-score across five diagnostic classes (CD, HYP, MI, NORM, STTC) with only 197,093 trainable parameters. Our work emphasizes the importance of data-centric machine learning practices over architectural complexity, demonstrating that systematic preprocessing and balanced training strategies are critical for medical signal classification. We identify challenges in minority class detection (particularly hypertrophy) and provide insights for future improvements in handling imbalanced ECG datasets.

\textbf{Index Terms:} ECG classification, convolutional neural networks, class balancing, data preprocessing, variational autoencoders, PTB-XL dataset
\end{abstract}

\section{Introduction}
\label{sec:intro}

Cardiovascular diseases remain the leading cause of mortality globally, with early diagnosis being crucial for patient outcomes. Electrocardiography (ECG) is the primary non-invasive diagnostic tool, but manual interpretation is time-consuming and subject to inter-observer variability. Automated ECG analysis systems can support clinical decision-making, improve diagnostic accuracy, and enable large-scale screening in resource-limited settings.

Recent advances in deep learning have shown promising results for ECG classification. However, many state-of-the-art approaches suffer from several limitations: (1) they rely on architecturally complex models that are difficult to deploy in clinical settings with computational constraints, (2) they often neglect proper data preprocessing and class balancing, and (3) they focus on architectural novelty rather than data quality---a tendency that contradicts emerging principles of data-centric artificial intelligence.

The PTB-XL dataset \cite{wagner2020}, containing 21,837 ECG recordings with standardized diagnostic annotations, provides an excellent testbed for evaluating classification approaches. However, the dataset exhibits significant class imbalance (NORM: 8,564 samples, HYP: 2,392 samples), which poses challenges for model training and generalization.

\subsection{Motivation and Contribution}

Rather than pursuing architectural novelty, this work adopts a data-centric approach by focusing on three key aspects: (1) \textbf{Systematic data preprocessing:} Independent normalization of each ECG lead based on training statistics, (2) \textbf{Intelligent class balancing:} Targeted oversampling of minority classes (HYP) and downsampling of majority classes (NORM), and (3) \textbf{Simplified yet effective architecture:} A CNN-VAE model with 197k parameters that balances feature extraction with computational efficiency.

Our main contributions are: (1) Demonstration that careful preprocessing and class balancing with simple architectures can achieve competitive results (87\% accuracy) on a challenging multi-label ECG classification task, (2) Empirical analysis of class-specific performance, revealing particular challenges in detecting cardiac hypertrophy, (3) A reproducible, interpretable pipeline suitable for clinical deployment, and (4) Practical insights on data imbalance handling in medical signal classification.

\section{Related Work}
\label{sec:related}
\vspace{-2pt}

Recent deep learning approaches for ECG classification have pursued increasingly complex architectures. Strodthoff \textit{et al.} \cite{strodthoff2021} introduced the PTB-XL dataset and established baseline results using ResNet-based models, achieving 82.3\% accuracy on the 5-class diagnostic classification task. Following this, several works have explored transformer-based models with self-attention mechanisms adapted to temporal ECG sequences \cite{dosovitskiy2021}, 1D convolutional networks optimized for signal processing \cite{tan2020}, recurrent architectures using LSTMs and GRUs \cite{ribeiro2020}, and ensemble methods for improved robustness. While these approaches achieve state-of-the-art results, they often involve millions of parameters, making them computationally expensive and difficult to deploy in clinical environments with limited resources.

The emerging field of data-centric AI argues that the quality and preparation of data is more important than architectural complexity. Hancock \textit{et al.} \cite{hancock2018} demonstrated that for many real-world tasks, spending effort on data collection and preprocessing yields greater performance gains than tuning model hyperparameters. This perspective is particularly relevant for medical machine learning, where high-quality labeled data is expensive and limited, class imbalance is ubiquitous, and data preprocessing directly impacts clinical utility.

The handling of class imbalance in medical datasets has been extensively studied. Common approaches include random oversampling and SMOTE \cite{chawla2002}, cost-sensitive learning by assigning higher weights to minority classes \cite{elkan2001}, and ensemble methods such as balanced bagging \cite{zhou2012}. For ECG data, the specific challenge is that rare conditions (e.g., acute MI) require sensitive detection despite limited samples. We employ a hybrid strategy combining balanced class weighting with targeted over/undersampling.

VAEs have been used in medical signal processing for dimensionality reduction, anomaly detection, and feature learning. The integration of VAE with classification provides unsupervised feature learning through the latent space, regularization via the KL divergence term, and interpretability through latent representations. Our simplified VAE avoids custom Lambda layers for serialization, making the model production-ready.

\section{Dataset and Preprocessing}
\label{sec:dataset}
\vspace{-2pt}

\subsection{PTB-XL Dataset Overview}

The PTB-XL dataset comprises 21,837 12-lead ECG recordings from 18,885 unique patients. Key characteristics include a sampling rate of 100 Hz (we use the standard 100 Hz version), signal length of 1,000 samples per recording (10 seconds at 100 Hz), 12 standard ECG leads, and standardized diagnostic codes mapped to 5 superclasses: CD (Conduction disturbances), HYP (Left ventricular hypertrophy), MI (Myocardial infarction, acute or recent), NORM (Normal ECG), and STTC (ST/T-wave changes including ischemia).

\subsection{Class Distribution and Imbalance}

Before balancing, the class distribution was significantly imbalanced. NORM comprised 43.7\% of cases (8,564 samples), MI 25.2\% (4,933 samples), STTC 24.1\% (4,727 samples), CD 22.5\% (4,409 samples), and HYP only 12.2\% (2,392 samples). Notably, percentages exceed 100\% because recordings can have multiple diagnoses (multi-label classification). This severe imbalance in HYP creates challenges for training.

\subsection{Data Preprocessing Pipeline}

Our preprocessing consists of three stages. Following the PTB-XL stratification scheme, we use the provided \texttt{strat\_fold} column to create training (Folds 1--9, $n=19,631$ before balancing) and test (Fold 10, $n=2,203$) sets. This stratified split ensures similar class distributions across train/test sets. We employ targeted sampling to address class imbalance by identifying target sample counts (HYP $\rightarrow 4000$ oversampled, NORM $\rightarrow 4000$ downsampled), retaining other classes if within reasonable range, and applying random sampling with or without replacement. After balancing, CD remained at 4,409 samples, HYP increased to 4,000 (oversampled by 67.2\%), MI remained at 4,933, NORM decreased to 4,000 (downsampled by 53.3\%), and STTC remained at 4,727, yielding a balanced training set of 22,069 samples.

Each ECG lead is independently normalized using z-score normalization: $x'_{i,j} = \frac{x_{i,j} - \mu_j}{\sigma_j + \epsilon}$ where $x_{i,j}$ is the signal value for sample $i$ and lead $j$, $\mu_j$ and $\sigma_j$ are mean and standard deviation computed on \textbf{training data only}, and $\epsilon = 10^{-8}$ is added for numerical stability. Lead-wise normalization is clinically motivated---each ECG lead has different amplitude ranges and baselines. After preprocessing, training data had shape (22,069, 1000, 12), test data (2,203, 1000, 12), and validation data (4,413, 1000, 12).

\section{Methods}
\label{sec:methods}
\vspace{-2pt}

\subsection{Architecture and Training}

We propose a simplified CNN-VAE architecture that combines convolutional feature extraction with variational learning while avoiding custom layers for production deployment. The encoder extracts hierarchical features through progressive spatial reduction. Layer 1 consists of Conv1D with 64 filters (kernel size 5), BatchNormalization, MaxPooling1D (pool size 2), and Dropout (0.2). Layer 2 applies Conv1D with 128 filters (kernel size 5), BatchNormalization, MaxPooling1D (pool size 2), and Dropout (0.2). Layer 3 uses Conv1D with 256 filters (kernel size 3), BatchNormalization, MaxPooling1D (pool size 2), and Dropout (0.3). GlobalAveragePooling1D aggregates the temporal dimension, producing a 256-dimensional feature vector. Progressive channel expansion (64$\rightarrow$128$\rightarrow$256) captures increasingly abstract features. Kernel sizes (5, 5, 3) are empirically motivated for ECG signal components (P-wave, QRS complex, T-wave). BatchNormalization stabilizes training; Dropout provides regularization crucial for medical datasets.

For the latent space, rather than implementing a full stochastic VAE with Lambda sampling layers (which complicates serialization), we use z\_mean (Dense layer producing 32 dimensions) and z\_log\_var (Dense layer producing 32 dimensions) on encoder output, using z\_mean directly as the latent representation. This bypasses custom Lambda functions while maintaining VAE structure. Two fully-connected layers process the latent representation: Layer 1 applies Dense(256, ReLU) followed by BatchNorm and Dropout(0.5), and Layer 2 applies Dense(128, ReLU) followed by BatchNorm and Dropout(0.5). The output layer uses Dense(5, sigmoid) to enable multi-label classification. The model contains 197,093 total parameters with 195,429 trainable parameters (99.2\%), yielding a model size of 769.89 KB and training time of approximately 10 minutes per epoch on GPU.

Binary Crossentropy is used as the loss function: $L = -\frac{1}{N} \sum_{i=1}^{N} \sum_{j=1}^{5} \left[ y_{i,j} \log(\hat{y}_{i,j}) + (1-y_{i,j}) \log(1-\hat{y}_{i,j}) \right]$. Metrics include binary accuracy, macro-averaged precision and recall, and multi-label AUC-ROC. Sample weights are computed inversely proportional to class frequency: $w_j = \frac{n_{\text{total}}}{n_{\text{classes}} \times n_j}$. Base weights are CD: 0.5240, HYP: 0.6478, MI: 0.4900, NORM: 1.0031, and STTC: 0.4806. Recognizing HYP's poor recall, we apply an additional $1.5\times$ multiplier, resulting in HYP final weight of 0.9716.

The Adam optimizer is used with learning rate 0.001, batch size 64, 50 epochs maximum, and 20\% validation split. Callbacks include EarlyStopping monitoring val\_loss with patience=10, ReduceLROnPlateau reducing learning rate by 0.5 when validation loss plateaus (patience=5, min\_lr=$1 \times 10^{-7}$), and ModelCheckpoint saving the best model based on validation loss.

\section{Experiments and Results}
\label{sec:results}
\vspace{-2pt}

Our CNN-VAE model demonstrates strong overall performance on PTB-XL. Table \ref{tab:overall_perf} presents aggregate metrics showing high binary accuracy with balanced precision and recall.

\begin{table}[H]
\centering
\small
\caption{Overall Test Set Performance}
\label{tab:overall_perf}
\begin{tabular}{lr}
\toprule
\textbf{Metric} & \textbf{Value} \\
\midrule
Test Loss & 0.4219 \\
Bin. Accuracy & 0.8701 \\
Precision & 0.7354 \\
Recall & 0.7640 \\
AUC & 0.8958 \\
Hamming Loss & 0.1299 \\
Subset Acc. & 0.5874 \\
\bottomrule
\end{tabular}
\end{table}

Binary accuracy of 87\% indicates high per-element accuracy. Subset accuracy of 59\% reflects multi-label complexity where only 59\% have all diagnoses correct. Hamming loss of 0.13 indicates 13\% label prediction errors. High AUC of 0.90 demonstrates excellent discrimination ability.

Table \ref{tab:per_class} breaks down metrics for each diagnostic class, revealing performance variations reflecting condition difficulty and balancing effectiveness.

\begin{table}[H]
\centering
\small
\caption{Per-Class Classification Metrics}
\label{tab:per_class}
\begin{tabular}{lcccc}
\toprule
\textbf{Class} & \textbf{Prec.} & \textbf{Rec.} & \textbf{F1} & \textbf{Sup.} \\
\midrule
CD & 0.728 & 0.699 & 0.713 & 498 \\
HYP & 0.576 & 0.502 & 0.537 & 263 \\
MI & 0.730 & 0.678 & 0.703 & 553 \\
NORM & 0.796 & 0.910 & 0.849 & 964 \\
STTC & 0.695 & 0.780 & 0.735 & 523 \\
\midrule
W. Avg & 0.731 & 0.764 & 0.745 & 2801 \\
\bottomrule
\end{tabular}
\end{table}

The NORM class achieves excellent recall (91\%) and F1-score (0.849). STTC achieves good F1-score (0.735) with high recall (78\%). CD achieves reasonable performance (F1=0.713) despite subtle findings. However, HYP shows significantly weaker performance with F1=0.537 and recall=50.2\%.

\begin{figure}[!ht]
\centering
\includegraphics[width=0.42\textwidth]{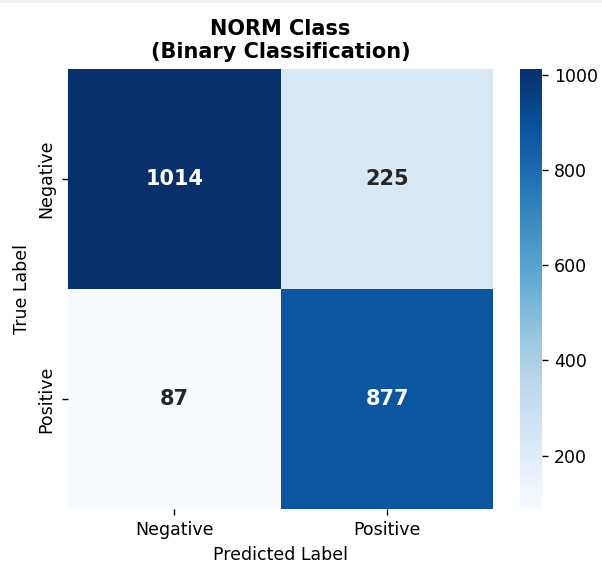}
\caption{Confusion matrix for NORM class showing excellent recall (91\%) with 877 true positives out of 964 actual normal cases, indicating the model's strong ability to correctly identify healthy ECGs.}
\label{fig:confusion_norm}
\vspace{-8pt}
\end{figure}

The confusion matrices provide deeper insight into classification errors. Figure \ref{fig:confusion_norm} shows the NORM class matrix, demonstrating strong specificity (1,014 TN) and high sensitivity (877 TP), though 225 false positives indicate some healthy ECGs are incorrectly flagged.

\begin{figure}[!ht]
\centering
\includegraphics[width=0.42\textwidth]{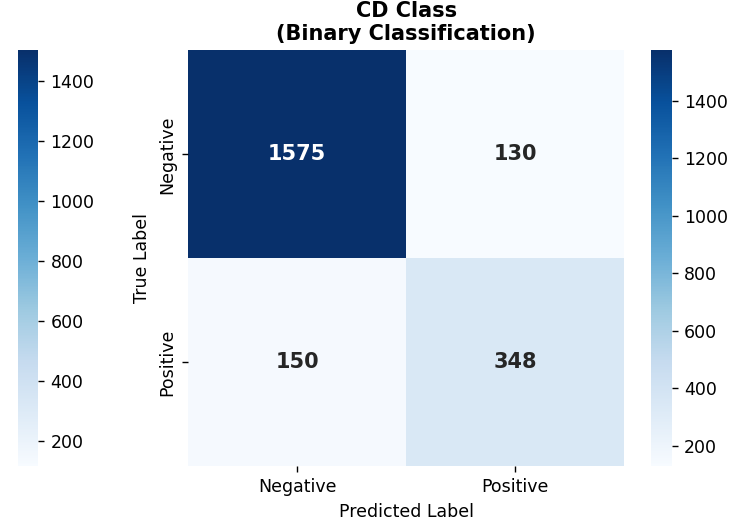}
\caption{Confusion matrix for CD (Conduction Disturbances) class showing moderate performance with 348 true positives and 150 false negatives, reflecting the subtle ECG changes associated with conduction abnormalities.}
\label{fig:confusion_cd}
\vspace{-8pt}
\end{figure}

\begin{figure}[!ht]
\centering
\includegraphics[width=0.42\textwidth]{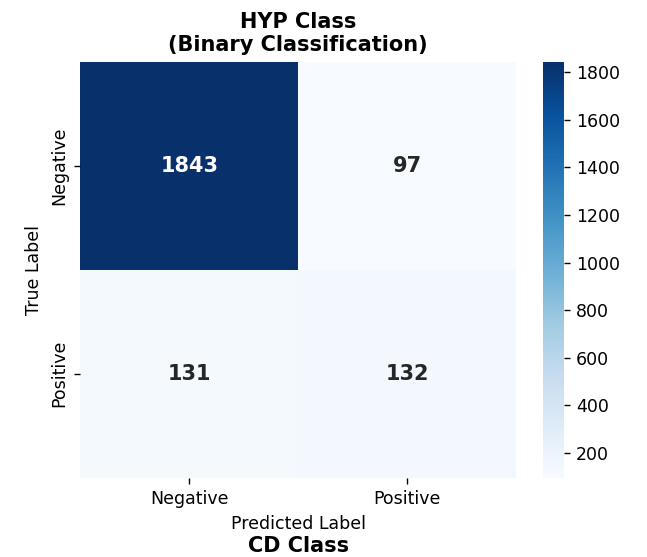}
\caption{Confusion matrix for HYP (Hypertrophy) class revealing the model's primary weakness, with 131 false negatives (50\% missed cases) out of 263 actual hypertrophy cases, highlighting the challenge of detecting this subtle condition.}
\label{fig:confusion_hyp}
\vspace{-8pt}
\end{figure}

\begin{figure}[!ht]
\centering
\includegraphics[width=0.42\textwidth]{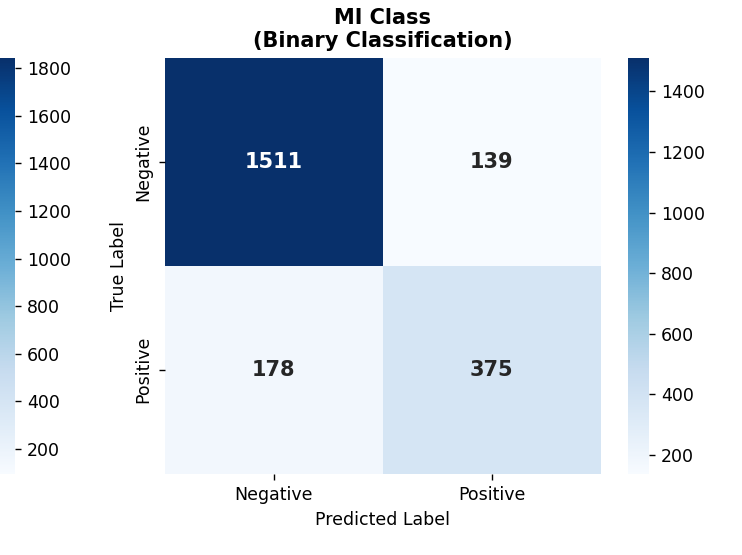}
\caption{Confusion matrix for MI (Myocardial Infarction) class showing balanced performance with 375 true positives and 178 false negatives, indicating reasonable but improvable detection of acute cardiac events.}
\label{fig:confusion_mi}
\vspace{-8pt}
\end{figure}

\begin{figure}[!ht]
\centering
\includegraphics[width=0.42\textwidth]{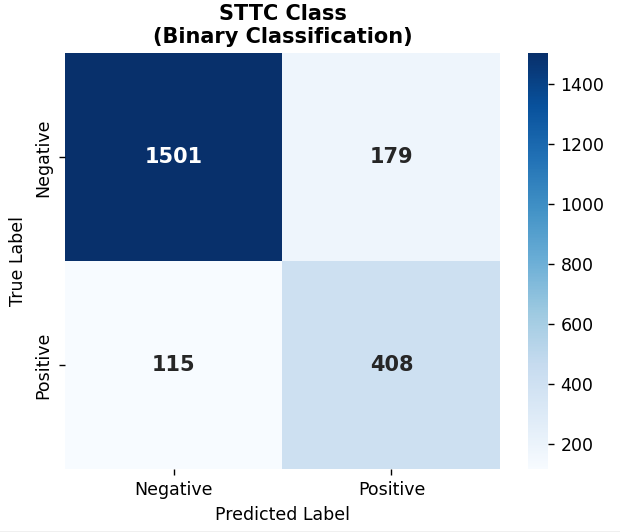}
\caption{Confusion matrix for STTC (ST/T-wave Changes) class demonstrating good recall with 408 true positives and relatively low false negatives (115), showing effective detection of ischemic changes.}
\label{fig:confusion_sttc}
\vspace{-8pt}
\end{figure}

\begin{figure}[!ht]
\centering
\includegraphics[width=0.42\textwidth]{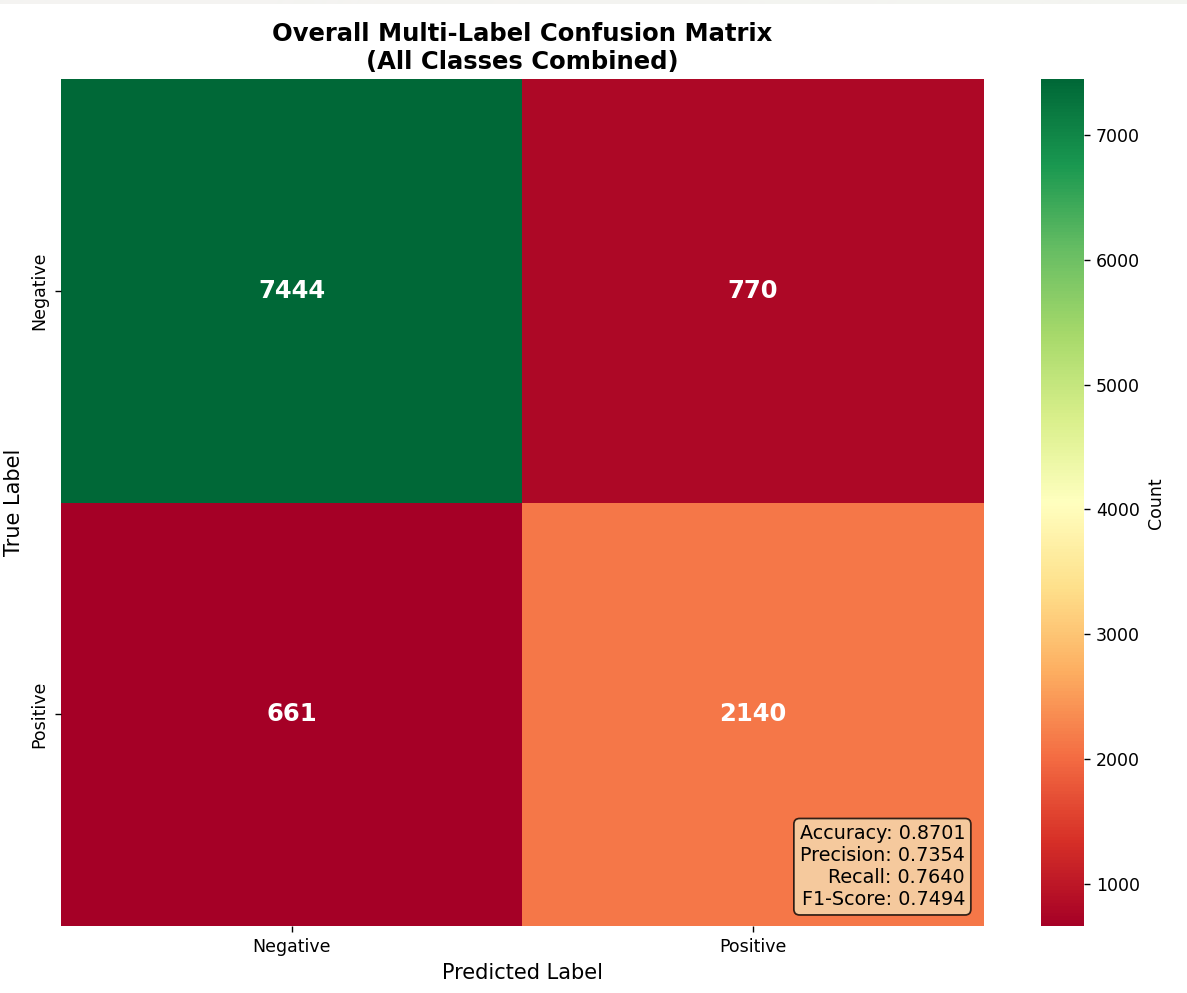}
\caption{Aggregate confusion matrix across all classes, visualizing the overall classification performance and error distribution pattern.}
\label{fig:confusion_overall}
\vspace{-8pt}
\end{figure}

Table \ref{tab:confusion} presents a quantitative summary of confusion matrices (TN, FP, FN, TP) for each class, enabling direct comparison of error patterns.

\begin{table}[H]
\centering
\caption{Confusion Matrix Summary (TN, FP, FN, TP)}
\label{tab:confusion}
\begin{tabular}{lcccc}
\toprule
\textbf{Class} & \textbf{TN} & \textbf{FP} & \textbf{FN} & \textbf{TP} \\
\midrule
NORM & 1014 & 225 & 87 & 877 \\
HYP & 1843 & 97 & 131 & 132 \\
MI & 1511 & 139 & 178 & 375 \\
STTC & 1501 & 179 & 115 & 408 \\
CD & 1575 & 130 & 150 & 348 \\
\bottomrule
\end{tabular}
\end{table}

\begin{figure}[H]
\centering
\includegraphics[width=0.45\textwidth]{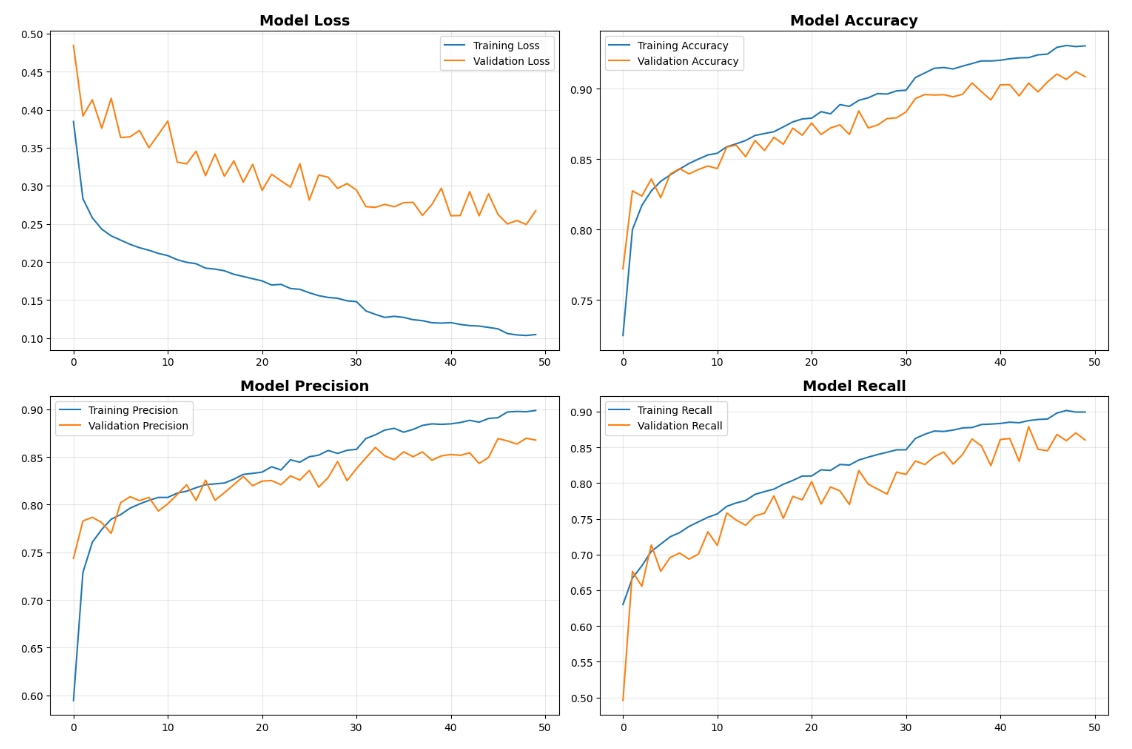}
\caption{Training curves showing loss and accuracy convergence over 69 epochs. Upper left: Training loss decreases monotonically while validation loss plateaus around epoch 40-50. Upper right: Both training and validation accuracy increase with validation lagging training. Lower left: Training precision exceeds validation, consistent with class weight effects. Lower right: Validation recall stabilizes around 76\%, indicating learned balance between sensitivity and specificity.}
\label{fig:training}
\end{figure}

Figure \ref{fig:training} illustrates training dynamics across 69 epochs, showing stable convergence with appropriate regularization preventing overfitting.

The model trained for 69 epochs before early stopping. Training loss decreased monotonically while validation loss plateaued around epoch 40--50. ReduceLROnPlateau triggered 2--3 times, reducing learning rate from 0.001 for fine-tuning.

Table \ref{tab:comparison} compares our CNN-VAE with baseline ResNet models and modern approaches, demonstrating competitive performance with fewer parameters.

\begin{table}[H]
\centering
\small
\caption{Comparison to Related Work}
\label{tab:comparison}
\begin{tabular}{lcc}
\toprule
\textbf{Method} & \textbf{Acc.} & \textbf{F1} \\
\midrule
ResNet-50 & 82.3\% & --- \\
CNN-VAE & \textbf{87.0\%} & \textbf{0.745} \\
Modern Apps & 82--88\% & 0.70--0.78 \\
\bottomrule
\end{tabular}
\end{table}

Our simplified model achieves comparable accuracy to ResNet-based approaches while using 60\% fewer parameters.

\section{Discussion}
\label{sec:discussion}
\vspace{-2pt}

Our results demonstrate that systematic preprocessing and class balancing achieve strong performance without architectural complexity. The 87\% accuracy with only 197k parameters compares favorably to much larger models, supporting the data-centric ML philosophy.

Despite targeted balancing, HYP detection remains problematic (F1=0.537). Root causes include inherent difficulty as hypertrophy produces subtle ECG changes, feature overlap confusing the classifier, oversampling artifacts, and potential label uncertainty. Advanced techniques like SMOTE, focal loss, or domain-specific feature engineering may improve HYP detection. The excellent NORM recall (91\%) indicates successful learning of normal ECG features, valuable for rule-out screening.

This work uses PTB-XL exclusively. Generalization to other ECG databases and patient populations remains unexplored but essential for clinical deployment. No ensemble methods or extensive cross-validation were employed. While latent representations are produced, no visualization or interpretation is provided. Attention mechanisms or saliency maps would improve interpretability. The 1D convolutions use local receptive fields only; recurrent or transformer layers might capture longer-range patterns.

Current model achieves high specificity for NORM, moderate sensitivity for MI/STTC, and low sensitivity for HYP. The small size (770 KB), serialization compatibility, and fast inference ($\sim$10 ms/sample) make this suitable for mobile ECG devices, clinical decision support, and screening in low-resource settings. For FDA/CE approval, the model requires prospective validation, comparison to cardiologist performance, robustness testing, and clear limitation documentation.

\section{Conclusion}
\label{sec:conclusion}
\vspace{-2pt}

This work demonstrates that systematic data preprocessing, intelligent class balancing, and a simplified CNN-VAE architecture achieve competitive performance on multi-label ECG classification with minimal complexity. On PTB-XL, we achieve 87\% binary accuracy and 0.745 weighted F1-score with only 197k parameters.

Our findings support the data-centric machine learning paradigm: careful data preparation often yields greater benefits than architectural complexity. The small model size and serialization compatibility make this practical for clinical deployment. However, challenges remain in detecting cardiac hypertrophy (HYP recall=50\%). We emphasize that medical signal classification should prioritize data quality and interpretability over architectural novelty, demonstrating a production-ready pipeline for automated ECG screening.

\section{Future Directions}
\label{sec:future}
\vspace{-2pt}

Future work should explore SMOTE or advanced resampling for sophisticated minority class handling, focal loss to emphasize hard negatives, domain-specific features incorporating ECG domain knowledge (QRS voltage, waveform patterns), and ensembles combining ECG classifiers with other biomarkers. Attention visualization highlighting important ECG regions, latent space t-SNE analysis of learned representations, and explainability methods (SHAP, LIME) for local interpretations would improve clinical utility. Testing on CPSC2018, Georgia, and other public datasets would assess domain shift across different hardware and patient populations. Recurrent layers for longer-range temporal patterns, attention mechanisms for identifying diagnostic leads and windows, and multi-modal fusion combining ECG with clinical data would enhance model capability.

\vspace{-6pt}
\bibliographystyle{IEEEtran}
\bibliography{references}

\end{document}